\documentclass[11pt,a4paper]{article}
\usepackage[hyperref]{acl2021}
\usepackage{times}
\usepackage{latexsym}
\usepackage{booktabs}
\usepackage{times}
\usepackage{bm}

\usepackage{multirow}
\usepackage{amsmath}
\usepackage{amssymb}
\usepackage{pifont}
\newcommand{\cmark}{\ding{51}}%
\newcommand{\xmark}{\ding{55}}%

\usepackage{microtype}

\aclfinalcopy 

\title{FST: the FAIR Speech Translation System for the IWSLT21 Multilingual Shared Task}

\author{{\parbox{0.75\linewidth}{\centering{Yun Tang*\quad\; Hongyu Gong*\quad\; Xian Li\quad\; Changhan Wang\quad\; Juan Pino\quad\; Holger Schwenk\quad\; Naman Goyal}}} \\
Facebook AI Research \\
  \texttt{\{yuntang,hygong,xianl,changhan,juancarabina,schwenk,naman\}@fb.com}
}
\date{}

\begin{document}
\maketitle
\begin{abstract}

In this paper, we describe our end-to-end multilingual speech translation system submitted to the IWSLT 2021 evaluation campaign on the Multilingual Speech Translation shared task.  Our system is built by leveraging transfer learning across modalities, tasks and languages. First, we leverage general-purpose multilingual modules pretrained with large amounts of unlabelled and labelled data. We further enable knowledge transfer from the text task to the speech task by training two tasks jointly. 
Finally, our multilingual model is finetuned on speech translation task-specific data to achieve the best translation results. 
Experimental results show our system outperforms the reported systems, including both end-to-end and cascaded based approaches, by a large margin. 
 In some translation directions, our speech translation results evaluated on the public Multilingual TEDx test set are even comparable with the ones from a strong text-to-text translation system, which uses the oracle speech transcripts as input. 
\end{abstract}
{\let\thefootnote\relax\footnote{{*Yun Tang and Hongyu Gong have equal contribution to this work.}}}

\section{Introduction} 


Multilingual speech translation~\cite{inaguma2019multilingual} enables translation from audio to text in multiple language directions with a single model. Similar to multilingual text translation, it is sample efficient as the model supports more languages. Furthermore, multilingual speech models can facilitate positive transfer across languages by learning a common representation space from speech inputs, typically either raw audio or filterbank features.

In this paper, we provide a description of our submission to the first multilingual speech translation task at IWSLT 2021. The task evaluates speech translations from Spanish (es), French (fr), Portuguese (pt) and Italian (it) into English (en) and Spanish (es). Among them, three translation directions (it-en, it-es and pt-es) are considered zero-shot with respect to the constrained track.
In addition, participants are encouraged to submit transcriptions for the relevant languages.

Our team, FAIR Speech Translation (FST), participated in the unconstrained track, where we submitted one primary system and four contrastive systems. 
We are interested in exploring the effectiveness of building a general-purpose multilingual multi-modality model.
We leverage large amounts of data, including unlabelled and labelled data from different modalities, to alleviate the data scarcity issue.  
We build the multilingual model to perform speech translation and speech recognition tasks for all evaluation directions. 
Our model leverages self-supervised pretraining on both the encoder and the decoder. The model is further improved by knowledge transferring from the text-to-text translation task to the speech-to-text translation task under the multitask learning framework. 
Finally, we finetune the model on parallel speech translation corpora as well as weakly aligned speech translation data through data mining to achieve the best result.

In \autoref{sec:data}, we described data sources and our method for speech translation data mining. Models and training methods are then described in \autoref{sec:method}. Finally, we present the results for the primary and contrastive systems in \autoref{sec:experiments}.

\section{Data}\label{sec:data}

Provided by the IWSLT 2021 shared task, the multilingual TEDx dataset collected from TED talks provides speech translations in $13$ directions \cite{salesky2021mtedx}. We focus on the seven competition directions in the shared task: es-en, fr-en, pt-en, it-en, fr-es, pt-es and it-es.

\begin{table*}[htbp!]
\caption{Audio Length in Hours of TEDx, CoVoST, EuroParl and Mined Data}
\label{tab:data_size}
\centering
\begin{tabular}{c|ccccccc}
\hline
 & es-en & fr-en & it-en & pt-en & fr-es & pt-es & it-es \\ \hline
TEDx & 163.7 & 119.9 & - & 134.2 & 85.5 & - &  - \\ \hline
CoVoST & 113.0 & 264.1 & 10.3 & 44.1 & - & - & - \\
EuroParl & 20.7 & 31.0 & 35.5 & 14.6 & 20.0 & 9.5 & 20.6 \\ \hline
Common Voice (mined data) & 52.7 & 39.6 & 12.8 & 9.6 & 18.7 & 4.4 & 6.6 \\
MLS (mined data) & 23.9 & 64.7 & 2.3 & - & 42.7 & 1.3 & 3.4 \\ \hline
\end{tabular}
\end{table*}

\subsection{Public data}

Besides TEDx dataset provided by the shared task, we also include two other public datasets, CoVoST and EuroParl, which provides parallel audio-text samples in some of the test directions of TEDx.
\begin{itemize}
    \item CoVoST~\cite{wang2020covost}. As a large scale dataset for multilingual speech translation, CoVoST contains translations from $11$ languages to English. We use its data in $5$ language directions \footnote{\{es, fr, it, pt, ru\}-en}.
    \item EuroParl \cite{iranzo2020europarl}. Collected from debates in European Parliment, EuroParl provides speech-to-text translations in $6$ European languages. Its data in $11$ language directions \footnote{es-\{en, fr, it, pt\}, fr-\{en, es, pt\}, it-\{en, es\}, pt-\{en, es\}, ru-en} is used in model training.
\end{itemize}

\subsection{Mined data}
We also mined additional speech-to-text data from unlabeled corpora. The audio corpora used in our experiments include Common Voice and Multilingual LibriSpeech (MLS).
\begin{itemize}
    \item Common Voice \cite{ardila2020common}. It is a massive collection of multilingual audios and their transcriptions in $29$ languages.
    \item MLS \cite{pratap2020mls}. It is a speech corpus collected from audiobooks of LibriVox in $8$ languages.
\end{itemize}

The text corpus used for mining is CCNet, which serves as the source of target translations  \cite{wenzek2020ccnet}. Collected from snapshots of CommonCrawl dataset, CCNet provides a large-scale and high-quality monolingual datasets. 

Since the audio corpora provide transcripts for audios, we could align source audios with target translations by finding the alignments between source transcripts and target texts. LASER alignment is applied for the crosslingual text alignment \cite{artetxe2019margin}. It generates sentence embeddings with a pre-trained multilingual text encoder \cite{schwenk2017learning}, and use them to measure the semantic similarity between sentences.

Table~\ref{tab:data_size} summarizes the statistics of the data used in our experiments. It reports the total length of audios in TEDx, CoVost and EuroParl datasets. Moreover, we include the statistics of mined speech from Common Voice and MLS. The mined data has an equivalent size to TEDx dataset in training directions. It also provides a good amount of speech data in zero-shot directions including it-en, pt-es and it-es.


\subsection{Text Data}\label{sec:data:text}
We use additional text data to train mBART model, which later is used to initialize our speech-to-text model. mBART model is first trained with monolingual text data from five languages\footnote{Five languages include en,es,fr,it and pt.} using self-supervised training. Then they are finetuned with parallel text data from seven evaluation directions as a multilingual text-to-text translation model.
The monolingual text data comes from the CC100 dataset~\cite{Conneau2020UnsupervisedCR} and the 
parallel text data are downloaded from OPUS~\cite{Tiedemann2012ParallelDT}.~\footnote{The following datasets are used: CommonCrawl,  OPUSBooks v1, CAPES v1, DGT v2019, ECB v1, ELRA-W0138 v1, ELRA-W0201 v1, ELRC 2682 v1, EMEA v3, EUbookshop v2, 
EuroPat v1, Europarl v8, GlobalVoices v2018q4, JRC-Acquis v3.0, JW300 v1b, Multi ParaCrawl v7.1, MultiUN v1, News-Commentary v14,  
QED v2.0a, SciELO v1, TED2013 v1.1, TED2020 v1, Tanzil v1, Tatoeba v2020-11-09, TildeMODEL v2018,UNPC v1.0, 
and UN v20090831, Wikipedia v1.0.}

\begin{table*}[htbp!]
\centering
\begin{tabular}{ c  | c  c  c  c | c  c  c | c} 
\toprule
   &	 \multicolumn{4}{|c|}{$\rightarrow$ en} &\multicolumn{3}{|c|}{$\rightarrow$ es} &   \\ 
  \cline{2-5}\cline{6-8}
  & es & fr & pt & it & fr & pt & it & Ave.\\
 \hline\hline
 MT  (M2M-100)~\cite{salesky2021mtedx} & 34.0 & 40.9 & 38.7 & 34.6 & 42.4 & 45.8 & 44.2 & 40.1 \\
 \hline
 \hline
 Cascaded System~\cite{salesky2021mtedx} & 21.5 & 25.3 & 22.3 & 21.9 & 26.9 & 26.3 & 28.4 & 24.7 \\
 Multilingual E2E~\cite{salesky2021mtedx} & 12.3 & 12.0 & 12.0 & 10.7 & 13.6 & 13.7 & 13.1 & 12.5 \\
 \hline
 \hline
 ST Baseline  & 27.8 & 32.4& 26.6 & 20.6  & 35.0 & 28.7 & 28.3  & 28.5 \\
 \hline
  XLSR-IPA   & 32.1 & 36.8& 35.1& 30.0& 38.3& 38.5 &37.5 & 35.5 \\
  XLSR-SPM   & 33.2 & 37.8& 35.0& 29.3& 39.5& 36.7 &35.3 & 35.3 \\
  VP100K-IPA  & 31.6 & 37.1& 35.3& 29.3& 38.2& 37.9 &37.1 & 35.2 \\
  \hline
  Ensemble (3 models) & 34.0& 38.7& 37.2& 30.9& 39.7& 40.4 &38.6 & 37.1 \\
 \hline
\end{tabular}
\caption{Main results on the public test set from the Multilingual TEDx Corpus~\cite{salesky2021mtedx}.}
\label{table:main_results}
\end{table*}

\section{Methods}\label{sec:method}
Our evaluation system is based on an encoder decoder model with the state-of-the-art Transformer architecture.
The submitted model is developed with a transfer learning approach~\cite{Li2020MultilingualST}, including three ingredients:
single-modality modules pretrained from self-supervised learning, multitask joint training, and task-specific fine-tuning.
The pretrained modules make use of a large amount of unlabeled data, joint training focuses on transferring knowledge from a relatively simple text-to-text task to a speech-to-text task, and the model is fine-tuned on speech-to-text translation task to boost in-task performance. 

\begin{table*}[ht!]
\centering
\begin{tabular}{ c  | c  c  c  c | c  c  c | c  c  c  c} 
\toprule

  &  \multicolumn{7}{|c}{BLEU }&\multicolumn{4}{|c}{}  \\
   \cline{2-8}
  &  \multicolumn{4}{|c|}{$\rightarrow$ en }&\multicolumn{3}{|c}{$\rightarrow$ es } &\multicolumn{4}{|c}{ WER}  \\
   \cline{2-8}\cline{9-12}
  & es & fr & pt & it & fr & pt & it & es & fr & it & pt\\
  \hline
 ST Baseline & 34.1 & 28.4 & 19.8 & 20.0 & 29.3 & 25.3 & 25.8 & 18.6 & 25.7 & 33.2 & 44.5  \\
  \hline
 XLSR-IPA   & 40.4  & 36.4 & 29.0 & 28.4 & 34.4 & 34.4 & 34.6 & 13.0 & 21.8 & 21.8 & 29.9 \\
  \hline
 Ensemble (3 models) & 42.2  & 38.7 & 31.0 & 29.4 & 36.5 & 38.2 & 37.3 & 11.2 & 18.7 & 19.6 & 27.4  \\
 \hline
\end{tabular}
\caption{Main results on the blind test set from the Multilingual TEDx Corpus.}
\label{table:main_results_blind}
\end{table*}

\subsection{Modality Dependent Pretraining}\label{sec:pre_training}
Our model leverages large amounts of unlabelled data from different modalities through two pretrained models: a wav2vec 2.0~\cite{Baevski2020wav2vecV2} and a multilingual BART (mBART)~\cite{Liu2020MultilingualDP}.

\noindent {\bf wav2vec 2.0} is a simple and powerful framework to learn high quality speech representation from unlabelled audio data. 
 Given the raw input audio samples, the model learns both latent representations and context representations through a contrastive task to distinguish true latent from distractors.
 Two multilingual wav2vec 2.0 models are explored during our development. One (``XLSR-53") is trained on 56K-hour speech in 53 languages~\cite{conneau2020unsupervised},
 and another (``VP-100K") is trained on 100K-hour speech in 23 languages~\cite{Wang2021VoxPopuliAL}.  
The pretrained wav2vec 2.0 models are used to initialize the speech encoder in the jointly trained model of the next stage. 

As will be discussed in our experiments, the two encoders are strong in different language directions. We ensemble models with XLSR-53 encoder and VP-100K encoder respectively to achieve the best performance.

\noindent {\bf mBART} is a sequence-to-sequence generative pretraining scheme, specifically a denoising autoencoder (DAE) to predict the original text from its noisy version such as random span masking and order permutation \cite{Liu2020MultilingualDP}. 
The model is pretrained with monolingual data and finetuned with parallel data as described in \autoref{sec:data:text}. 
The encoder and decoder in mBART model are used to initialize the  encoder and  decoder in the joint trained model of the second stage.

Previous study~\cite{Tang2020AGM} shows that it makes the knowledge transfer from the text-to-text task to speech-to-text task easier by representing the input text as its pronunciation form, i.e., the phoneme sequence.
We also investigate representing the input text as its pronunciation forms rather than sentencepiece tokens during our development.  We choose International Phonetic Alphabet (IPA) as input text representation since it can be shared across different languages. espeak\footnote{http://espeak.sourceforge.net/index.html} is used to convert the text word into IPA phonemes. 

\begin{table*}[htb!]
\centering
\begin{tabular}{c|ccc|cccc|ccc|c} 
\toprule
\multirow{2}{*}{} &  \multicolumn{3}{c|}{Data} &  \multicolumn{4}{c|}{$\rightarrow$ en} & \multicolumn{3}{c|}{$\rightarrow$ es} & \multirow{2}{*}{Ave.} \\ \cline{1-11} 
 & ASR & Public & Mined & es & fr & pt & it & fr & pt & it &  \\ \hline\hline
M0 & \xmark & \xmark & \xmark & 22.3 & 26.7 & 21.7 & 5.9 & 28.2 & 23.6 & 8.4 & 19.5 \\
M1  & \cmark & \xmark & \xmark & 24.2 & 29.1 & 26.3 & 18.1 & 31.7 & 28.9 & 27.3 & 26.5  \\
M2 & \cmark & \cmark & \xmark  & 25.2 & 30.8 & 26.9 & 19.2 & 32.5 & 29.4 & 28.1 & 27.4  \\
M3 (ST Baseline)  & \cmark & \cmark & \cmark & 27.8 & 32.4 & 26.6 & 20.6 & 35.0 & 28.7 & 28.3 & 28.5  \\ \hline
\end{tabular}
\caption{Ablation studies of training data (Public: CoVoST and EuroParl, Mined: mined data from Common Voice, MLS and CCMatrix, ASR: ASR data in mTEDx). The results are BLEU scores on TEDx test set. The models considered here are built upon pretrained XLSR-53 encoder and mbart decoder, and they do not have joint training. The speeach translation data from mTEDx is used by all models. }
\label{tab:data_analysis}
\end{table*}

\subsection{Multitask Joint Training}
In the second stage, we choose to optimize the speech-to-text translation model along with a text-to-text translation  model. 
Two encoders are used to process text input and speech input respectively. The speech encoder is with the large wav2vec 2.0 configuration. The feature extractor and the bottom 12 transformer layers in the context extractor are initialized with the corresponding parameters from the pretrained wav2vec 2.0 model in \autoref{sec:pre_training}. The top 12 transformer layers in the speech encoder are shared with the text encoder. They are initialized with the pretrained mBART encoder~\cite{Tang2021IST}. 
An adaptor~\cite{Li2020MultilingualST}, which consists of 3 1-D convolution layers with stride 2 to achieve 8x down-sampling of speech encoder outputs, is placed between the last non-shared speech encoder layer and the first shared speech text encoder layer.    The decoder is shared by two tasks and initialized with the pretrained mBART decoder. 

Two techniques~\cite{Tang2021IST}: cross attentive regularization (CAR) and online knowledge distillation (online KD), are employed to enhance the knowledge transferring.  Text input data comes from the corresponding transcripts in the speech translation dataset. Due to time limits, we don't use extra parallel text data to enhance the performance.

\subsection{Speech only Finetuning}
In the last stage, the model is fine-tuned in the speech-to-text translation task with speech input only. The text encoder is dropped and no text input data is used.  
\section{Experiments} \label{sec:experiments}
\subsection{Experimental Setting}
 Both wav2vec 2.0 model and mBART model are trained with the large configuration. There are 24 transformer layers in the wav2vec 2.0 model, and 12 transformer layers in both mBART encoder and decoder.  We build the mBART model with a target vocabulary of 64,000 SentencePiece~\cite{Kudo2018SentencePieceAS} tokens, which are shared among all 6 evaluation languages\footnote{In our evaluation, the new mBART model achieves comparable results as the public available mBART model, which is with 250k vocabulary, but with much smaller memory footprint.}.  For the mBART model with IPA phoneme input, the vocabulary size is 516 which includes phoneme variants with ``\_'' attached to denote the word leading phoneme.  
 
A language id symbol ``$\langle$LID$\rangle$'' is used as the initial token to predict the sentence. Speech recognition task is treated as the same as the speech translation task but with the source speech language id symbol.

The primary system results submitted are from an ensemble system with three models.  All three models are trained with 3-stage optimization discussed in \autoref{sec:method} with different initialization models. The first one is initialized with ``XLSR-53'' wav2vec model and IPA mBART model (``XLSR-IPA''). Compared with the first model, the second model chooses sentence piece mBART model (``XLSR-SPM'') while the third one is initialized with ``VP-100K'' wav2vec model (``VP100K-IPA'') \footnote{We will release the training and evaluation recipe under https://github.com/pytorch/fairseq/tree/master/examples/speech\\\_text\_joint\_to\_text}. 

We use 8 V100 GPUs for each model during the jointly training and fine-tuning stages.
It takes approximate five days to jointly train the models for 15 epochs and another two days for speech only fine tuning.
The last 5 checkpoints are averaged for inference with beam size 5.

\begin{table*}[htb!]
\centering
\begin{tabular}{c|cc|cccc|ccc|c} 
\toprule
\multirow{2}{*}{} &  \multicolumn{2}{c|}{Train} &  \multicolumn{4}{c|}{$\rightarrow$ en} & \multicolumn{3}{c|}{$\rightarrow$ es} & \multirow{2}{*}{Ave.} \\ \cline{1-10} 
 & JT & FT & es & fr & pt & it & fr & pt & it &  \\ \hline\hline
M3  & \xmark & \cmark & 27.8 & 32.4 & 26.6 & 20.6 & 35.0 & 28.7 & 28.3 & 28.5 \\
M4  & \cmark &  \xmark & 32.3 & 36.6 & 33.8 & 28.4 & 38.3 & 35.9 & 35.7 & 34.4 \\
M5  & \cmark & \cmark &  33.2 & 37.8& 35.0& 29.3& 39.5& 36.7 &35.3 & 35.3 \\ \hline
\end{tabular}
\caption{Ablation studies of training approaches (JT: joint training of text and speech translation, FT: finetuning a trained model on TEDx data in speech translation). The results are BLEU scores reported on TEDx test set. The models considered here are built upon pretrained XLSR-SPM encoder and mbart decoder. They are trained with the combination of TEDx including the ASR portion, public data as well as mined data.}
\label{tab:train_analysis}
\end{table*}

\begin{table*}[htb!]
\centering
\begin{tabular}{c|c|cccc|ccc|c} 
\toprule
\multirow{2}{*}{} & \multirow{2}{*}{Encoder} &  \multicolumn{4}{c|}{$\rightarrow$ en} & \multicolumn{3}{c|}{$\rightarrow$ es} & \multirow{2}{*}{Ave.} \\ \cline{3-9} 
 & & es & fr & pt & it & fr & pt & it &  \\ \hline\hline
M4 & XLSR-SPM & 32.3 & 36.6 & 33.8 & 28.4 & 38.3 & 35.9 & 35.7 & 34.4  \\ 
M6 & VP100K-SPM & 30.5 & 35.6 & 33.7 & 28.5 & 36.9 & 36.9 & 36.2 & 34.0 \\ \hline
\end{tabular}
\caption{Ablation studies of different encoders. BLEU scores are reported on TEDx test set. Models are jointly trained on all data, but they are not further finetuned on speech translation.}
\label{tab:encoder_analysis}
\end{table*}
To provide a deep insight into the factors affecting translation performance, we conduct ablation studies on different model configurations. 

\subsection{Main Results}

We summarize our main results at \autoref{table:main_results}. The first row presents results from a large text-to-text translation system (M2M-100)~\cite{Fan2020BeyondEM} using oracle speech transcripts as input~\cite{salesky2021mtedx}. The second two rows list best results from the cascaded system and multilingual end to end system from literature~\cite{salesky2021mtedx}. The fourth row to eighth row are results from our systems. 
The fourth row presents our multilingual baseline, which is initialized with pretrained wav2vec 2.0 model (``XLSR-53'') for encoder and mBART model (``SPM'') for decoder. 
The model is fine-tuned with Multilingual TEDx data, public data and mined data listed in \autoref{sec:data}. No joint training is applied.  
``XLSR-IPA'', ``XLSR-SPM'' and ``VP100K-IPA'' from row 5 to 8 are results from the 3 best systems we built.  Compared with the baseline in the third row,  these three systems have an extra step to co-train with the text-to-text translation task. 

It is clear that we create a very strong baseline (row 4) with the help from the large amounts of speech/text training data.
In comparison to the previous reported cascaded system (row 2) or multilingual end-to-end system (row 3), the results are 3.8 and 16.0 BLEU scores higher on average.

Row 5 to 8 provide evaluation results from our 3 best single models built with single-modality based pre-training, multitask joint training and task-specific fine-tuning.  They are built with different pre-training data or input text representations. 
Compared with the baseline in row 4, another 6.7 $\sim$ 7.0 BLEU improvement are observed. 
IPA phoneme based text representation gives slight gain compared with text units separated with SentencePiece model (``XLSR-IPA'' v.s. ``XLSR-SPM''), which is smaller than we expected. 
We hypothesis that it is due to the imperfect text to phoneme conversion for different languages.  The difference due to different pre-training data is also small that there are only 0.3 BLEU in average when the speech pre-training is changed (``XLSR-IPA'' v.s. ``VP100K-IPA'').

The ensemble of three models achieves the best performance with a $1.6$ BLEU improvement over the best single model.
It indicates those three models are complementary to each other, though they give similar BLEU scores in our test.  The results are even close to the ones from the strong text-to-text translation system (M2M-100 in row 1), which takes speech transcript as translation input. Our primary system achieves the same BLEU score as the text-to-text translation system on translation direction "es-en" and the average BLEU score gap from 7 directions is 3.0.

The corresponding blind test results are reported in \autoref{table:main_results_blind}. Similar to our observation in~\autoref{table:main_results}, the model trained with the 3-stage approach significantly improves the translation accuracy compared with the baseline. The ensemble system outperforms other systems in all speech translation directions as well as the speech recognition tasks.

\subsection{Analysis}

\textbf{Data}. Table~\ref{tab:data_analysis} compares models trained with different sets of data. Additional data is shown to improve the translation performance. In our multitask training, we combine the text-to-text and speech-to-text translation tasks together. We don't include ASR task as separated task, instead we treat ASR task as a special translation direction. It shows ASR data is helpful for speech translation, especially for translation directions with small amount of speech training data (``it-en'' and ``it-es''). On average, we observe a significant gain of $7.0$ BLEU from the comparison between M0 and M1 . 

When we continue adding public datasets including CoVost and EuroParl to the training set, M2 has an average improvement of $0.9$ BLEU over M1. The mined data brings another $1.1$ BLEU gain as we compare M3 and M2.


\textbf{Training}. Different training approaches are compared in Table~\ref{tab:train_analysis}. We observe significant gains brought by joint training of text and speech translation. Compared against M3, M4 with joint training demonstrates an improvement of $5.9$ BLEU over $7$ language directions. When the jointly trained model is further finetuned with speech translation data, an extra gain of $0.9$ is achieved as we compare M5 against M4.


\textbf{Encoder}. We compare XLSR-53 and VP-100K encoder in Table~\ref{tab:encoder_analysis}. XLSR-53 is strong at encoding audios in Spanish and French, achieving BLEU gains of $1.8$ and $1.0$ in es-en and fr-en respectively. VP100k encoder outperforms XLSR-53 in  pt-es and it-es directions with gains of $1.0$ and $0.5$ respectively. This can be explained by the fact that VP100K encoder is trained on more Portuguese and Italian Speech.







\section{Conclusion}
 In this work, we described our multilingual end-to-end speech translation system submitted to IWSLT 2021. We leverage the large amount of training data from different domains and modalities to improve the speech translation performance. We adopt a progressive approach to build the model with three stages. Compared with our strong baseline, the proposed system achieves 8.6 BLEU score improvement, which also outperforms other reported systems, including both end-to-end and cascaded based, by a large margin.

\bibliographystyle{acl_natbib}
\bibliography{acl2021}

\appendix

\end{document}